%% file: neurips_2025.tex
\documentclass{article}

\usepackage[nonatbib, final]{neurips_2025}

\usepackage[utf8]{inputenc} 
\usepackage[T1]{fontenc}    
\usepackage{hyperref}       
\usepackage{url}            
\usepackage{booktabs}       
\usepackage{amsfonts}       
\usepackage{nicefrac}       
\usepackage{microtype}      
\usepackage{xcolor}         
\usepackage{graphics}
\usepackage[pdftex]{graphicx}

\hypersetup{hidelinks} 
\usepackage{multirow}
\usepackage{mathtools,amssymb}
\usepackage{enumitem}
\usepackage{float}
\usepackage[vlined,ruled,linesnumbered]{algorithm2e}
\usepackage{tabularx}
\usepackage[small,compact]{titlesec}
\usepackage{subcaption}
\usepackage{tablefootnote}
\newcolumntype{Y}{>{\centering\arraybackslash}X}

\title{SimulMEGA: MoE Routers are Advanced Policy Makers for Simultaneous Speech Translation}

\author{%
  Chenyang Le \\
  \texttt{nethermanpro@sjtu.edu.cn}
  \And 
  Bing Han \\
  \texttt{hanbing97@sjtu.edu.cn}
  \And 
  Jinshun Li \\
  \texttt{muzi-jingshun@sjtu.edu.cn}
  \And 
  Songyong Chen \\
  \texttt{chensy30@sjtu.edu.cn}
  \And 
  Yanmin Qian \thanks{Corresponding Author} \\
  \texttt{yanminqian@sjtu.edu.cn}
  \AND
  \texttt{Auditory Cognition and Computational Acoustics Lab} \\
\texttt{MoE Key Lab of Artificial Intelligence, AI Institute}\\
\texttt{School of Computer Science, Shanghai Jiao Tong University, Shanghai, China}
}

\begin{document}

\maketitle

\begin{abstract}
Simultaneous Speech Translation (SimulST) enables real-time cross-lingual communication by jointly optimizing speech recognition and machine translation under strict latency constraints. Existing systems struggle to balance translation quality, latency, and semantic coherence, particularly in multilingual many-to-many scenarios where divergent read/write policies hinder unified strategy learning. In this paper, we present \textbf{SimulMEGA}(\textbf{Simul}taneous Generation by \textbf{M}ixture-of-\textbf{E}xperts \textbf{GA}ting), an unsupervised policy learning framework that combines prefix-based training with a Mixture-of-Experts refiner to learn effective read/write decisions in an implicit manner, without adding inference-time overhead. Our design requires only minimal modifications to standard transformer architectures and generalizes across both speech-to-text and text-to-speech streaming tasks. Through comprehensive evaluation on six language pairs, our 500 M-parameter speech-to-text model outperforms the Seamless baseline, achieving under 7\% BLEU degradation at 1.5~s average lag and under 3\% at 3~s. We further demonstrate SimulMEGA’s versatility by extending it to streaming TTS with a unidirectional backbone, yielding superior latency–quality trade-offs. \footnote{The code can be found in \url{https://github.com/nethermanpro/simulmega}.}
\end{abstract}

\section{Introduction}
Simultaneous Speech Translation (SimulST) addresses the critical need for real-time cross-lingual communication by jointly optimizing speech recognition and machine translation under strict latency constraints. Unlike conventional offline systems that process complete utterances, SimulST operates on streaming audio input, incrementally generating translations while simultaneously decoding ongoing speech - a capability mirroring human interpreters' cognitive processing. This technology enables transformative applications in international diplomacy, live media localization, and low-latency dialogue systems. However, it faces fundamental challenges in reconciling three competing objectives: translation quality, computational latency, and semantic coherence, exacerbated by fragmented acoustic patterns and partial contextual dependencies.

Recent progress combines adaptive segmentation strategies\cite{ma2020SimulMT, dong2022Learning, zhangEndtoEndSimultaneousSpeech2023} with streaming architectures like RNN-T transducers\cite{xue2022LargeScale, deng2024LabelSynchronous, zhao2025Textless} and Monotonic Attention mechanisms\cite{raffel2017Online, ma2019Monotonic}, alongside policy-based decision frameworks\cite{guo2023Learning, yin2023Language, chen2024DivergenceGuided, zhang2024StreamSpeecha, zhao2024PsFuture, guo2025Large}. Despite these advances, current systems struggle to achieve human-like efficiency-accuracy trade-offs across diverse linguistic contexts and acoustic conditions. A particularly underexplored challenge lies in multilingual many-to-many translation, where divergent read/write policies across language pairs complicate the learning of unified operational strategies.

This paper introduces SimulMEGA, an unsupervised policy learning framework that synergizes prefix-based training with Mixture-of-Experts (MoE) mechanisms to address these limitations. Our approach enables high-performance multilingual SimulST through three key innovations: (1) An auxiliary MoE refiner module enabling implicit policy learning without inference-time overhead, (2) Minimal architectural modifications to standard transformers for broad applicability, and (3) A unified framework supporting both speech-to-text and text-to-speech streaming tasks. Experimental results demonstrate state-of-the-art performance across six languages, with our 500M-parameter speech-to-text model outperforming SeamlessM4T\cite{communication2023Seamless} by achieving <7\% BLEU degradation at 1.5s average lagging (AL) and <3\% at 3s AL. What's more, we extended our methods to streaming text-to-speech (TTS) task via CosyVoice2's unidirectional backbone\cite{du2024CosyVoice}, the system achieves superior latency control compared to conventional interleaved streaming approaches.

Our principal contributions are as follows.

\begin{itemize}[leftmargin=0.5cm]
\item \textbf{Introduction of SimulMEGA}: An unsupervised policy learning framework that combines prefix-based training with a Mixture-of-Experts (MoE) refiner module, enabling efficient and effective simultaneous speech translation without adding inference-time overhead.
\item \textbf{Multilingual Advancement}: SimulMEGA achieves SOTA performance for quality-latency tradeoffs in many-to-many translation, demonstrating cross-lingual robustness through comprehensive 6-language evaluation.
\item \textbf{Universal Streaming Framework}: SimulMEGA supports both speech-to-text and text-to-speech streaming tasks within a single framework, making it broadly applicable and easily integrable into existing models for multilingual SimulST tasks.
\item \textbf{General Conversational System}: The integrated solution combining low-latency speech-to-text and text-to-speech conversion, validated through real-world dialog interpretation scenarios. 
\end{itemize}

\input{relatedworks}

\input{method}

\section{Experiments and Results}

\subsection{Experiment Settings}
\paragraph{Task \& Data}We collect data and train on many-to-many speech-to-text translation across six popular languages: EN, ZH, DE, ES, FR, and IT. We collect various open-source speech recognition datasets covering the six languages, including  LibriSpeech~\cite{panayotov2015librispeech}, Multilingual Librispeech(MLS) ~\cite{pratap2020mls}, VoxPopuli ~\cite{wang2021voxpopuli}, Common Voice\cite{ardila2019common}, WenetSpeech\cite{zhang2022wenetspeech}, KeSpeech\cite{tang2021kespeech} and Emilia~\cite{emilia}. Then we create pseudo translation labels for these data by translating the transcription into different languages through a cloud text-to-text translation API. The dataset consists of approximately 100K hours of training data. For TTS experiment, we use a subset of ST training set that only contains Chinese and English data due to the language compatibility of the CosyVoise 2.  

\paragraph{Model} Our offline base model for ST is derived from the whisper medium model. For inference efficiency, we transformed the encoder into a streaming encoder and pruned half of the layers in the decoder. This results in an encoder with 20 chunk-AR blocks and 4 NAR blocks, and a decoder with 12 layers. Offline base model for TTS is CosyVoice 2 as mentioned in section~\ref{sec:tts}. In both the ST and TTS experiment, the MoE refiner consists of $N_\mathrm{refiner} = 6$ layers, with the hidden size matching that of the base model.

\paragraph{Training} Each experiment is trained in FP16 with 8 Nvidia H800 GPUs. In stage 1 offline training of the ST experiment, to retain the capability of the whisper encoder,  we employ Low-Rank Adaptation(LoRA)\cite{hu2021lora}  ($\alpha = 64$ ) at the chunk-AR blocks of the encoder. We train the offline model for 1M steps, which takes around 1 week. In stage 2 training, the chunk-AR blocks are frozen. The router and MoE Refiner module are randomly initialized. The stage 2 training takes about 2 days. We use AdamW optimizer\cite{loshchilov2017decoupled} and linear learning rate scheduler with 5000 steps of warmup. The maximum learning rate is 1e-4 in stage 1 training and 1e-5 in stage 2 training.

\paragraph{Evaluation} In S2TT evaluation, we employ case-sensitive BLEU\cite{papineni2002Bleu} with punctuation\footnote{https://github.com/mjpost/sacrebleu\cite{post-2018-call}} as our primary quality metric and average lagging (AL)\footnote{Code is borrowed from https://github.com/facebookresearch/SimulEval\cite{simuleval2020}} \cite{ma2019STACL} for latency assessment. Our evaluation spans two benchmark datasets: CoVoST2 \cite{wang2021covost} and Fleurs \cite{conneau2023fleurs}. For CoVoST2, we report averaged scores across 5 non-English to English (X-EN) translation pairs and 2 English to non-English (EN-X) pairs. The Fleurs dataset enables comprehensive many-to-many evaluation through its parallel data across six languages, yielding results for all 30 possible language pair combinations. We present separate averages for 5 X-EN pairs, 5 EN-X pairs, and 30 X-X pairs.

For TTS evaluation, we combine LibriSpeech-PC \textit{test-clean} \cite{panayotov2015librispeech, meister2023librispeech} and Seed-TTS \textit{test-zh} \cite{anastassiou2024seedtts} with CoVoST2 data. Our analysis includes recognition word error rate (WER), speaker similarity (SIM), and text-unit-to-speech-unit average lagging(AL). In S2ST evaluation on CoVoST2, we measure ASR-BLEU against speech-to-speech-unit AL. For English metrics, we utilize Whisper-Large-V3, while Mandarin evaluations employ Paraformer \cite{gao2022paraformer}. And for SIM, we employ a speaker verification model based on WavLM-large \cite{chen2022wavlm} to extract speaker embeddings. These embeddings are then used to compute the cosine similarity between synthesized speech and ground truth speech.

\paragraph{Baselines} In simultaneous S2TT, our method is compared against the Seamless model family \cite{communication2023Seamless} and four custom baselines: Wait-K \cite{ma2019STACL}, Dig-SST \cite{chen2024DivergenceGuided}, EDATT \cite{papi2023Attention} and AlignATT \cite{papi2023AlignAtt}, all implemented using our offline base model architecture. For streaming TTS comparisons, we benchmark against the native streaming implementation in CosyVoice2. In simultaneous S2ST evaluation, Seamless serves as our primary comparator. Both the Seamless framework and our proposed system incorporate voice preservation capabilities. We also compare against two publicly available models, StreamSpeech\cite{zhang2024StreamSpeecha} and NAST-S2S\cite{ma2024Nonautoregressive}, whose parameter size and data size are smaller.

\subsection{Main Results}

\paragraph{Offline Multilingual S2TT Performance}
The results are summarized in Table~\ref{tab:offline}. Leveraging high-quality pseudo-labeled data, our offline base model achieves performance comparable to SeamlessM4T large-v2 across our in-domain six languages, despite the latter having three times the parameter count. The Seamless models, constrained by an English-centric data distribution, slightly underperform ours in the FLEURS many-to-many evaluation. For simultaneous models, we report BLEU score degradation (in brackets) relative to their respective base models. Our analysis reveals that SimulMEGA exhibits minimal degradation, ranging from \textbf{0.3\%} to \textbf{3\%}, while Seamless suffers a more significant \textbf{7\%} drop. Consequently, SimulMEGA outperforms Seamless in 4 out of the 5 evaluated scenarios.

\begin{table*}[tb]
\centering
\caption{The offline BLEU score(\%) of different models on CoVoST2 and Fleurs testset, where all results are based on greedy search.  Only parameters involved in the S2TT inference are calculated. The value in the bracket denotes the performance degradation compared to the offline base model.}
\label{tab:offline}
\begin{tabular}{lcccccc}
\bottomrule
 \multirow{2}{*}{Models}  & \multirow{2}{*}{Param} & \multicolumn{2}{c}{CoVoST2} & \multicolumn{3}{c}{Fleurs}\\
 \cmidrule(lr){3-4}  \cmidrule(lr){5-7}
 &  & X-EN& EN-X& X-EN&  EN-X& X-X\\ \hline
  \multicolumn{7}{c}{Offline Models}\\
  SeamlessM4T Medium& 821M & 34.4& 35.9& 25.6& 27.1& 16.1\\
   SeamlessM4T Large-v2& 1.5B & \textbf{38.3}& \textbf{40.8}& \textbf{29.8}& 31.5& 19.6\\
 S2T Base (Ours)& 561M & 37.0& 38.9& 26.4 & \textbf{32.4}&\textbf{25.1}\\
 \hline
   \multicolumn{7}{c}{Simultaneous Models}\\
 Seamless-S2T& 2.0B& 35.3\tiny{(-7.8\%)}& 37.6\tiny{(-7.8\%)}& \textbf{28.0}\tiny{(-6.0\%)}&  28.9\tiny{(-8.3\%)}& 18.1\tiny{(-7.7\%)}\\
  SimulMEGA-S2T (Ours)& 561M& \textbf{36.9}\tiny(\textbf{-0.3\%})& \textbf{38.5}\tiny{(\textbf{-1.0\%})}& 26.3\tiny{(\textbf{-0.4\%})}&  \textbf{31.4}\tiny{(\textbf{-3.1\%})}& \textbf{24.7}\tiny{(\textbf{-1.7\%})}\\
 \bottomrule
\end{tabular}
\end{table*}

\begin{figure}[htp]
    \centering
    \includegraphics[width=1.0\linewidth]{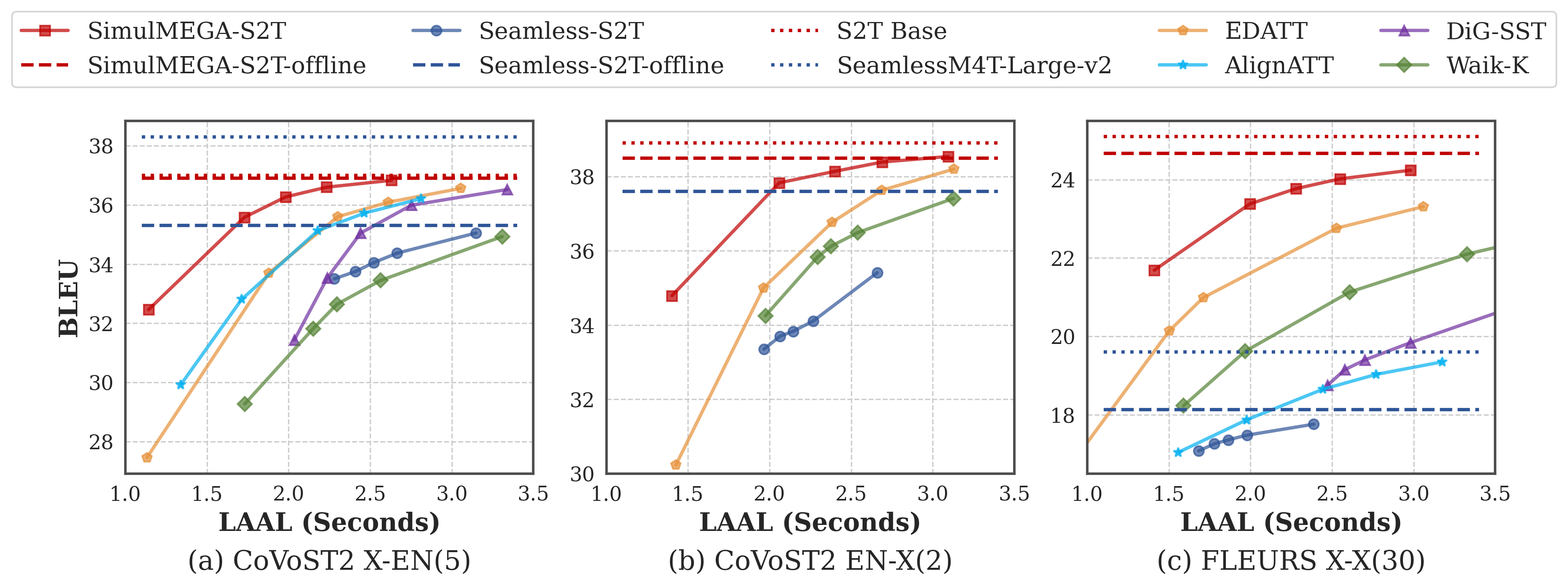}
    \caption{The multilingual simultaneous speech-to-text translation quality (BLEU) against the latency metrics (LAAL) on different testsets.  The reported BLEU and LAAL are the average of all language splits in the testsets(5 in CoVoST X-EN, 2 in CoVoST2 EN-X and 30 in Fleurs X-X)}
    \label{fig:simul}
\end{figure}

\paragraph{Simultaneous Multilingual S2TT Performance}
To evaluate generalizability across language pairs, we apply the same threshold configuration to all methods under identical settings. As illustrated in Figure~\ref{fig:simul}, SimulMEGA demonstrates robust performance across diverse language pairs and outperforms baseline methods in all three evaluation scenarios. Compared to the offline base model, SimulMEGA exhibits minimal degradation—\textbf{3\%}, \textbf{3\%}, and \textbf{5\%} degradation at an \textbf{AL of ~2 seconds}—while Seamless suffers significantly higher degradation (\textbf{12\%}, \textbf{17\%}, and \textbf{9\%}). Notably, SimulMEGA achieves low-latency performance with \textbf{<7\%} degradation at \textbf{1.5 seconds AL} and nears offline model quality (\textbf{<3\%} degradation) at \textbf{3 seconds AL}.

In contrast, Seamless, despite its self-learning design, struggles with a substantial performance gap due to limitations in multi-head monotonic attention(MMA). The static wait-K strategy incurs high latency and degradation across all settings. Meanwhile, DiG-SST, ED-ATT and AlignATT exhibits instability, failing on specific language pairs. In our experiment ED-ATT is most stable among baselines but still shows more than 0.5s additional LAAL compare to SimulMEGA under same quality.

\begin{figure}[htp]
    \centering
    \includegraphics[width=\linewidth]{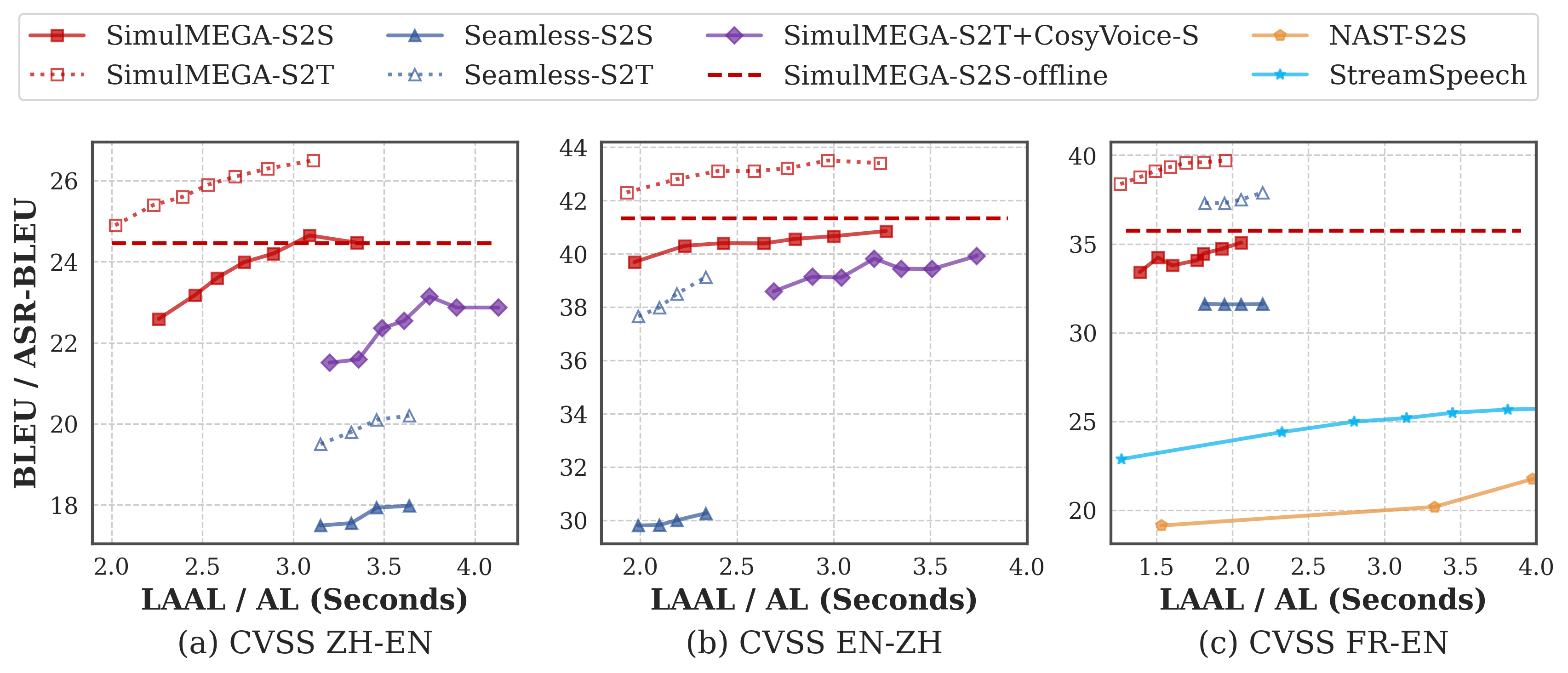}
    \caption{Bi-directional Simultaneous S2TT and S2ST Result between Mandarin and English in CoVoST2 test set. The translation quality metric is BLEU for S2TT and ASR-BLEU for S2ST. We use the same threshold group between S2ST and S2TT. Results of NAST-S2S and StreamSpeech are taken from original paper.}
    \label{fig:s2st}
\end{figure}

\paragraph{Streaming TTS \& Simultaneous S2ST} Our experiments on LibriSpeech and SeedTTS\_zh evaluate SimulMEGA-TTS under extreme streaming conditions, processing one text unit at a time. For the CoVoST2 cross-lingual setting, the system receives incremental text chunks generated by SimulMEGA-S2T. We compare against CosyVoice2-S operating at its default text-to-speech ratio. As shown in Table~\ref{tab:tts}, SimulMEGA-TTS demonstrates comparable speaker similarity to both CosyVoice2 variants, attributable to their shared flow model. While showing marginally lower alignment latency (AL) than CosyVoice systems, SimulMEGA-TTS maintains equivalent speech intelligibility (WER) to the original CosyVoice2. In cross-lingual evaluation, SimulMEGA-TTS surpasses Seamless by 10 SIM points while achieving more than 40\% lower WER. Figure~\ref{fig:s2st} presents our simultaneous speech-to-speech translation (S2ST) results. SimulMEGA-S2S maintains tight latency control, adding less than 200 ms AL compared to speech-to-text translation (S2TT). The quality degradation from S2TT to S2ST shows a 7\% (ZH-EN) and 6\% (EN-ZH) BLEU reduction for our system, lower than Seamless's 10\% and 20\% reductions, respectively. Furthermore, SimulMEGA-S2S outperforms synthesizing by CosyVoice2's streaming mode in both output quality and latency metrics.

\begin{table}[tb]
    \centering
    \caption{TTS evaluation results. ZS denotes zero-shot mode (prompt added in AR stage), and CL denotes cross-lingual mode (prompt not added in AR stage). ALs are calculated by the number of input text tokens. We use the generated text as the WER label for CoVoST2 evaluation. }
    \label{tab:tts}
    \resizebox{\linewidth}{!}{
        \begin{tabular}{lcccccccccc}
        \hline
        \multirow{2}{*}{} & \multicolumn{3}{c}{LibirSpeech} & \multicolumn{3}{c}{SeedTTS\_zh} & \multicolumn{2}{c}{CoVoST2\_zh-en} &  \multicolumn{2}{c}{CoVoST2\_en-zh}\\
         \cmidrule(lr){2-4} \cmidrule(lr){5-7} \cmidrule(lr){8-9} \cmidrule(lr){10-11}
        & WER & SIM & AL & WER & SIM & AL & WER & SIM  & WER&SIM\\
        \hline
        CosyVoice2-ZS& 2.44 & 0.658 & 22.3 & 1.62 & 0.757 & 23.1 & - & -  & -&-\\
        CosyVoice2-S-ZS\footnotemark[2]& 5.31& 0.651 & 18.5& 7.98& 0.760& 20.9& -& -& -&-\\
 CosyVoice2-S-CL\footnotemark[2]& 3.13& 0.535& 8.4& 4.33& 0.707& 8.4& 19.94& 0.388& 7.74&0.352 \\
        Seamless & - & - & - & - & - & - & 20.05& 0.292& 18.86 &0.301  \\
        SimulMEGA-TTS & 2.54 & 0.661 & 1.2& 1.90 & 0.755 & 0.5& 12.26& 0.391& 4.51&0.405\\
        \hline
    \end{tabular}
    }
\end{table}

\subsection{Ablation Study}

We conduct ablation experiments on three key designs in SimulMEGA: 1) pre-Sigmoid noise scale, 2) router score normalization, and 3) training losses. For these ablations, we employ the same base model and conduct stage-2 simultaneous training as in the main experiment. Training is performed on a subset of the main dataset containing only the Common Voice English data, with evaluation on the CoVoST2 EN-ZH test set.\footnotetext[2]{Using the officially released code and checkpoint (Commit fbab274), CosyVoice 2 exhibits hallucination issues in streaming mode, leading to increased WER and deviations from reported results.}

\begin{figure}[htp]
    \centering
    \includegraphics[width=1.0\linewidth]{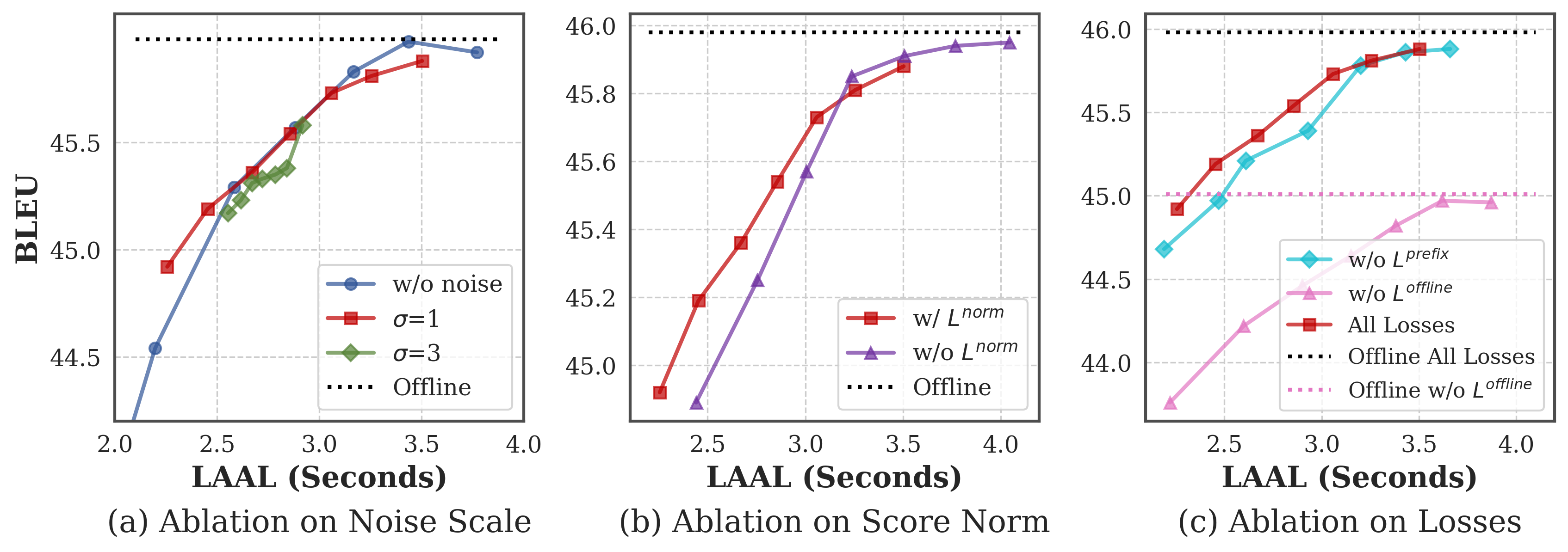}
    \caption{Ablation Studies. The threshold for all ablation experiment evaluation is 0.8, 0.7, ..., 0.2 from left to right.}
    \label{fig:ablation}
\end{figure}

\paragraph{Pre-Sigmoid Noise}
We examine three noise settings: 1) No noise, 2) $\sigma=1$, and 3) $\sigma=3$. Figure~\ref{fig:ablation}(a) indicates that noise magnitude has minimal effect on average performance. However, increasing noise leads to more deterministic router scores, thus limiting the flexibility of the latency-performance curve. Conversely, the absence of noise results in an overly wide score range and unstable low-latency performance. A moderate noise level ($\sigma=1$) achieves an optimal balance between a manageable dynamic range and stable translation performance.

\paragraph{Score Normalization}
Figure~\ref{fig:ablation}(b) shows that without normalization, router scores are disproportionately concentrated between thresholds 0.5 and 0.8, complicating threshold selection across different tasks or language pairs. Incorporating normalization yields a smoother latency-performance curve, reflecting a more balanced score distribution within the [0.2, 0.8] range, thereby facilitating easier and more consistent threshold tuning.

\paragraph{Training Loss}
We investigate the effects of removing either $\mathcal{L}^{\mathrm{offline}}$ or $\mathcal{L}^{\mathrm{prefix}}$ during simultaneous training. Results presented in Figure~\ref{fig:ablation}(c) demonstrate that removing $\mathcal{L}^{\mathrm{offline}}$ reduces overall performance by approximately 1 BLEU point, highlighting the necessity of offline loss to maintain simultaneous translation quality. Conversely, omitting $\mathcal{L}^{\mathrm{prefix}}$ leads to negligible performance degradation, suggesting its optional nature. We hypothesize that the generalization to prefix translation tasks provided by $\mathcal{L}^{\mathrm{offline}}$ and $\mathcal{L}^{\mathrm{refiner}}$ mitigates the need for truncation-specific training.

\section{Conclusion and Discussion}
SimulMEGA addresses multilingual SimulST challenges through an unsupervised policy learning framework integrating prefix-based training and Mixture-of-Experts refinement. By enabling minimal architectural modifications to standard transformers, it achieves state-of-the-art quality-latency tradeoffs across six languages while supporting unified speech/text streaming. The integrated system advances real-time cross-lingual communication with miniature BLEU degradation at low latency and robust conversational performance, establishing a versatile foundation for low-latency multimodal translation systems.

\paragraph{Limitation} Though SimulMEGA-S2S is a strong simultaneous S2ST system, it suffers from the drawback shared by cascaded systems: inconsistent text tokens(between Whisper tokens in S2TT and Qwen2 tokens in TTS), extra latency(around 0.1 seconds) and so on. What's more, SimulMEGA-TTS only supports two languages by now, which hinders its utility. In the future, we plan to add more language support, as well as integrate into a more unified end-to-end system. Additionally, currently SimulMEGA only support 30 seconds of maximum input duration. Therefore, it still rely on an VAD model to cut input stream into less than 30s. In the future we will explore continuously generation without segmentation by sliding window or history selection. 

\paragraph{Broader Impacts} This work proposes new streaming techniques and is dedicated to providing a faithful translation of the original speech. However, with the voice cloning capability of SimulMEGA-TTS, the system may be subject to some misuse. 

\section{Acknowledgment}
This work was supported by China STI 2030-Major Projects under Grant No. 2021ZD0201500

\medskip

\small
\bibliographystyle{IEEEtran}
\bibliography{mybib}

\input{appendix}

\end{document}

%% file: relatedworks.tex
\section{Related Works}

\begin{figure}
  \centering
  \includegraphics[width=1.0\linewidth]{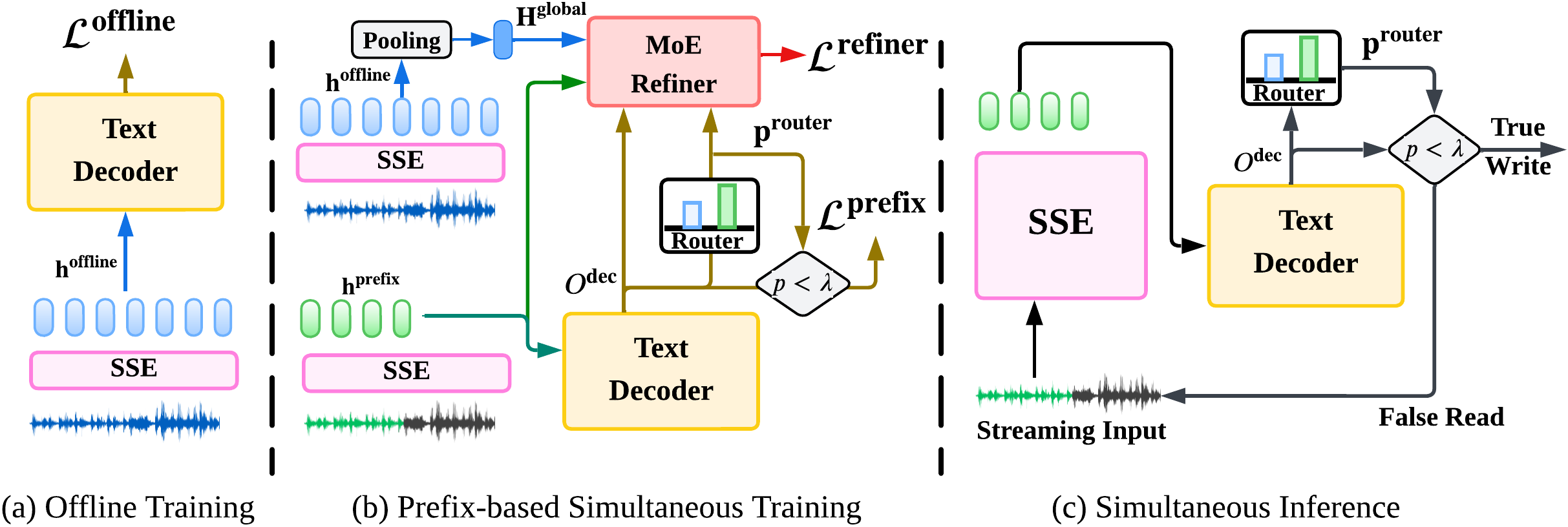}
  \caption{Overview of the training and inference paradigm of SimulMEGA.  SimulMEGA is composed of a Streaming Speech Encoder, a text decoder, an MoE routing Gate(Router), and an MoE refiner. In the first stage, the model is pre-trained on $\mathcal{L}^\mathrm{offline}$. In the second stage, the model is trained on a combination of $\mathcal{L}_\mathrm{offline}$, $\mathcal{L}^\mathrm{prefix}$ and $\mathcal{L}^\mathrm{refiner}$.  SSE denotes Streaming Speech Encoder. }\label{fig:overview}
\end{figure}

\paragraph{SimulST} Early works in SimulST focus on finding the input segmentation boundary and then applying a variant of the wait-k policy\cite{ma2020SimulMT, zeng2021RealTranS, dong2022Learning, zhangEndtoEndSimultaneousSpeech2023}. Later work seeks to find a more sophisticated dynamic policy. \cite{zhang2022InformationTransportbased} designs a learnable matrix between source and target for an integrate-and-fire(IF) policy, and \cite{papi2023Attention, papi2023AlignAtt} use attention score for a similar approach. \cite{chen2024DivergenceGuided} proposes a policy based on divergence between the prefix and full input. \cite{zhang2024StreamSpeecha} employs an extra CTC\cite{graves2006Connectionista} head for policy control. Another approach includes using an inherently streaming architecture like monotonic attention\cite{raffel2017Online, arivazhagan2019Monotonic, ma2019Monotonic, ma2023Efficient}, transducer\cite{xue2022LargeScale, deng2024LabelSynchronous, zhao2025Textless} and Hidden Markov\cite{zhang2022Hidden}. \cite{ma2024Nonautoregressive} explores non-autoregresive generation for SimulST.

\paragraph{Streaming TTS} Many works on TTS claim to be streamable\cite{lajszczakBASETTSLessons2024, dang2024LiveSpeech}, yet they are not optimized for fine-grained(chunk or token level) streaming and suffer from degraded quality due to a lack of future information. \cite{dekel2023Speaka} adapts a streaming model from an offline model via restricted attention. \cite{sheng2025SyncSpeech} uses a fixed number of text tokens for look-ahead. \cite{dang2024ZeroShot} processes each text chunk in a separate AR loop with look-ahead. \cite{du2024CosyVoice} interleaves text and speech tokens at a fixed ratio, which potentially incurs larger latency.

%% file: method.tex
\section{SimulMEGA: Simultaneous Translation via Mixture-of-Experts Routing}

In this section, we introduce SimulMEGA (\textbf{Simul}taneous Generation by \textbf{M}ixture-of-\textbf{E}xperts \textbf{GA}ting), an innovative framework that effectively converts an offline autoregressive model into a simultaneous system with minimal computational overhead and negligible performance degradation. 

The core principle of simultaneous training lies in enabling the model to autonomously determine whether the currently available input suffices for generating the next output token. Previous approaches typically rely on either artificial policy signals, which constrain the model's self-learning capacity, or architectural modifications that compromise generation quality. In contrast, our framework leverages a Mixture-of-Experts (MoE) architecture to achieve unsupervised policy learning while maintaining an identical structure to the offline model, thereby preserving translation performance. Our approach employs a prefix-based training strategy that closely mimics real-world streaming inference scenarios.

\subsection{Architecture Overview}
Figure \ref{fig:overview} illustrates the overall architecture of our SimulMEGA framework, which consists of four key components: (1) a streaming speech encoder, (2) a text decoder, (3) a global routing gate, and (4) a Mixture-of-Experts (MoE) refiner module.

The streaming speech encoder and text decoder adopt the Transformer architecture, maintaining compatibility with conventional speech-to-text translation (S2TT) systems. As illustrated in Figure~\ref{fig:structure}(a), the encoder combines chunk-wise autoregressive (Chunk-AR) blocks with non-autoregressive (NAR) blocks in a hybrid design. The Chunk-AR blocks optimize inference efficiency through a cached key-value mechanism, while the NAR blocks preserve translation quality by capturing global context. To handle streaming inputs lacking an explicit end-of-sequence (EOS) token, we prepend a learnable end-of-stream (EoSt) flag (a binary embedding) to the NAR blocks’ input, signaling whether the current chunk terminates the audio stream. The text decoder follows a standard autoregressive Transformer architecture.

The core innovation lies in the routing gate and MoE refiner, which enable unsupervised learning of simultaneous translation policies without architectural modifications. The MoE refiner follows a transformer-like architecture and shares language model head with the text decoder, which combines prefix information and global information and predicts the target translation sequence. It is only activated during training, introducing zero additional computational overhead during inference.

\subsection{Unsupervised Policy Learning.}
\begin{figure}
  \centering
  \includegraphics[width=0.8\linewidth]{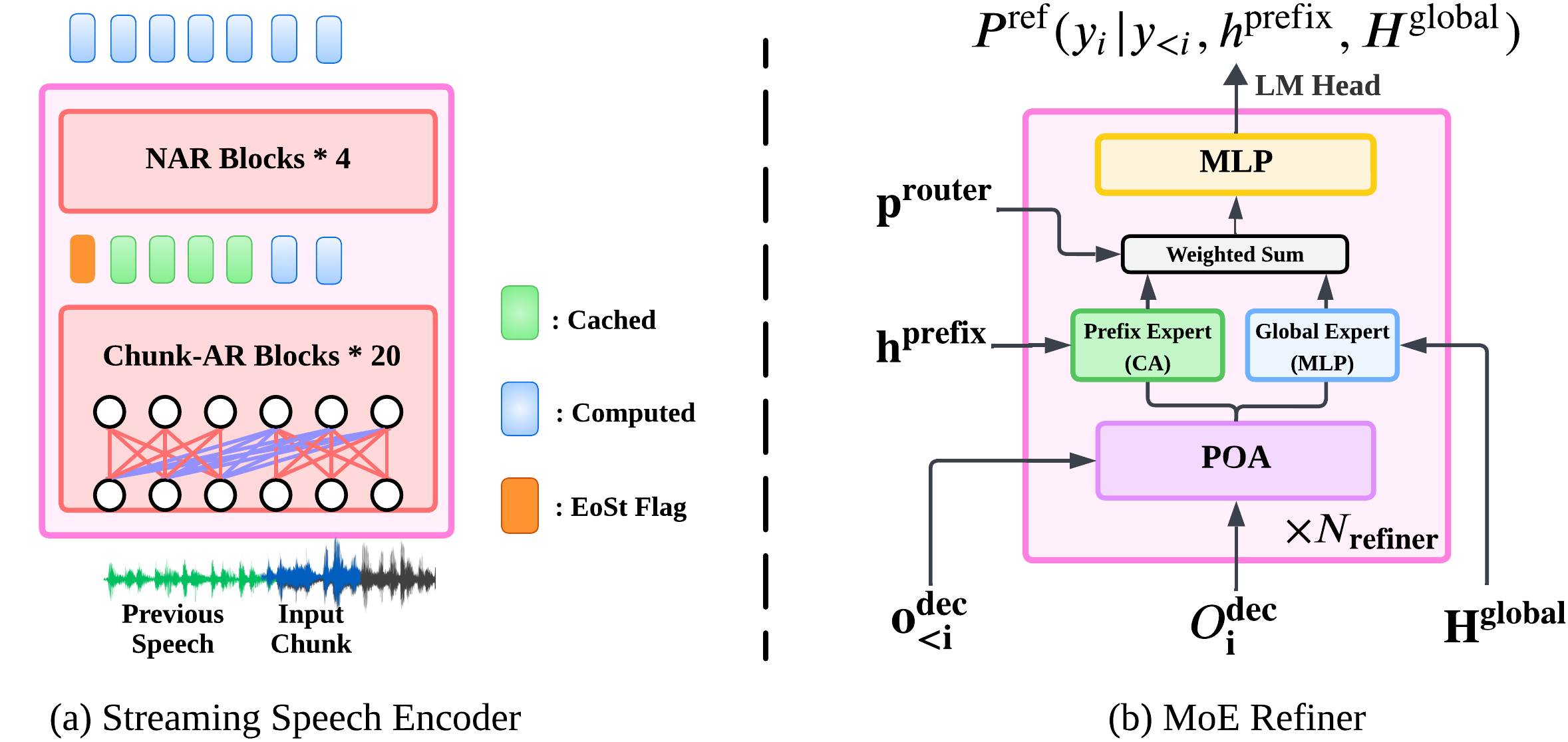}
  \caption{(a) Structure and inference example of the streaming speech encoder of SimulMEGA. It comprises 20 chunkwise autoregressive (Chunk-AR) blocks and 4 non-autoregressive(NAR) blocks. After each read, the Chunk-AR blocks only compute the new chunk while the NAR blocks recompute the whole sequence. An End-of-Stream (EoSt) flag is given before NAR blocks. (B) The structure of the MoE Refiner, in which the gate decides over the mixture proportion of the Prefix Expert and the Global Expert. This proportion reflects the model's confidence in the prefix sequence, leading to a natural read/write policy. The self-attention module is replaced by a previous output attention(POA) module to prevent global information leakage.}\label{fig:structure}
\end{figure}

This section introduces the Mixture of Experts (MoE) refiner module, whose architecture is shown in Figure~\ref{fig:structure}(b). The module comprises $N_\text{refiner}$ blocks, each containing two specialized experts: a prefix expert ($E_p$) and a global expert ($E_g$). The merging weights of these experts implicitly define a policy that determines whether the current input prefix contains sufficient information for generating the target token. Besides the dual-experts module, each block also contains a previous-output attention module (similar to self-attention) and an MLP module, which is analogous to a standard transformer decoder block.  

\paragraph{Global Routing Gate} For decision consistency, each refiner layer employs a shared global gate implemented as a two-layer MLP with a Sigmoid head. The gate projects the text decoder's final hidden state ($O^\mathrm{dec}$) into a scalar value $p\in [0,1]$, which determines the expert weights: $P^\mathrm{Router}_{E_g}=p$ and $P^\mathrm{Router}_{E_p}=1-p$.

\paragraph{Dual-Expert Architecture} As shown in Figure~\ref{fig:overview}(b), during training, the encoder processes both the complete offline speech and a randomly truncated prefix, producing corresponding hidden states $h^\mathrm{offline}$ and $h^\mathrm{prefix}$. Then the gate and the dual-expert architecture will decide which hidden states to use. While the router typically favors the more informative $h^\mathrm{offline}$, we introduce an information bottleneck to balance this preference. Specifically, we apply temporal mean-pooling to $h^\mathrm{offline}$, yielding a condensed global embedding $H^\mathrm{global}$ for $E_g$. $E_g$ is a two-layer MLP module whose calculation is as follows:
\begin{equation}
    \mathbf{\widehat{x^\mathrm{in}_{i}}} = \mathrm{LayerNorm}\left(\mathbf{x^\mathrm{in}_{i}} \right)
\end{equation}
\begin{equation}
\mathbf{x_{i}^{E_g}} = \mathbf{W}^\mathrm{E_g}_\mathrm{out} \cdot \mathrm{ReLU}\left(\mathbf{W}^\mathrm{E_g}_\mathrm{in} \cdot \left[\mathbf{\widehat{x^\mathrm{in}_{i}}}; \mathbf{H^\mathrm{global}}\right] \right) 
\end{equation}
where $x^\mathrm{in}_{i}$ denotes the $i$-th position's input, $[\cdot;\cdot]$ represents vector concatenation,  $\mathbf{W}^\mathrm{E_g}_\mathrm{in/out}$ are learnable projection weights (biases omitted). The prefix expert $E_p$ employs a standard cross-attention:
\[
\mathbf{x_{i}^{E_p}} = \mathbf{W}^\mathrm{E_p}_\mathrm{O} \cdot
\mathrm{MHA}\left(
\mathbf{W}^\mathrm{E_p}_\mathrm{Q} \mathbf{\widehat{x^\mathrm{in}_{i}}}, \;
\mathbf{W}^\mathrm{E_p}_\mathrm{K} h^\mathrm{prefix}, \;
\mathbf{W}^\mathrm{E_p}_\mathrm{V} h^\mathrm{prefix}
\right)
\]
Where MHA denotes multi-head attention. The final output combines both experts' contributions through a gated residual connection:
\[
\mathbf{x^\mathrm{out}_{i}} = \mathbf{x^\mathrm{in}_{i}} + 
P^\mathrm{Router}_{i, E_g} \cdot \mathbf{x_{i}^{E_g}} +
 P^\mathrm{Router}_{i, E_p} \cdot \mathbf{x_{i}^{E_p}}
\]

\paragraph{Global Information Leakage} In standard transformer architectures, the self-attention mechanism inherently leaks global information across the sequence. This poses a critical issue for our design: even when the gate assigns $P^\mathrm{Router}_{E_g}=0$ for a certain position, the hidden state at that position may still access global context via self-attention, thereby undermining the intended expert specialization. To enforce strict isolation of global information, we replace self-attention with a previous-output attention mechanism. Instead of attending to hidden states within the same layer, each position only attends to the decoder’s prior outputs($o_{<i}^{dec}$), effectively preventing unintended information flow while preserving sequential dependencies.

\paragraph{Training Protocol} SimulMEGA leverages a pre-trained offline S2TT model as its foundation. The training process consists of two stages:
\begin{enumerate}[leftmargin=0.5cm]
    \item \textbf{Offline Pretraining:} We first train the model with the standard offline S2TT objective $\mathcal{L}^\mathrm{offline}$(incorporating streaming chunk masks in the encoder) until convergence, as illustrated in Figure~\ref{fig:overview}(a).
  \item \textbf{Simultaneous Training:} We then introduce two additional loss functions to enable simultaneous capabilities as illustrated in Figure~\ref{fig:overview}(b):
      \begin{equation}
        \mathcal{L}^\mathrm{refiner} = -\sum_i \mathrm{log}~p^\mathrm{ref}(y_i|y_{<i}, h^\mathrm{prefix}, H^\mathrm{global})
    \end{equation}
    \vspace*{-1\baselineskip} 
    \begin{equation}
        \mathcal{L}^\mathrm{prefix} = -\sum_{i : p_i < \lambda} \log p^\mathrm{dec}(y_i \mid y_{<i}, h^\mathrm{prefix})
    \end{equation}
    where $p^\mathrm{ref}$ and $p^\mathrm{dec}$ are the output distribution of the MoE refiner and the text decoder. $\lambda$ is a pre-defined hyperparameter that restricts the losses to the confident position.$\mathcal{L}^\mathrm{refiner}$ learns the read/write policy while $\mathcal{L}^\mathrm{prefix}$ strengthens the prefix-based translation capability. In this stage, the total loss is the weighted sum of the above three losses:
\begin{equation}
    \mathcal{L}^\mathrm{simul} = \mathcal{L}^\mathrm{offline} + w_\mathrm{r} \cdot \mathcal{L}^\mathrm{refiner} + w_\mathrm{p} \cdot \mathcal{L}^\mathrm{prefix}
\end{equation}
In our experiment, we set $w_\mathrm{r} = w_\mathrm{p} = 0.2$ to prioritize the offline training. 
\end{enumerate}

\paragraph{Inference Policy} During inference, SimulMEGA employs a straightforward threshold-based policy:
\begin{equation}
\text{Action} = \begin{cases}
\text{Write} & \text{if } p_{t,i} < \lambda \\
\text{Read} & \text{otherwise}
\end{cases}
\end{equation}
where $p_{t,i}$ is the gating score at input time $t$ and target position $i$.

\subsection{Score Distribution Control}
The raw output scores from the routing gate lack inherent interpretability, as they emerge solely from neural network optimization through gradient descent. These scores exhibit inconsistent statistical properties, both in mean and variance, between different tasks, language pairs, and training data. This variability poses significant challenges during inference, necessitating task-specific or language-specific threshold tuning that fundamentally compromises the generalizability of our approach. To overcome this limitation, we introduce the following techniques:

\paragraph{Score Normalization}
We introduce a heuristic that aligns the average gating score with the relative information content between prefix and global contexts. Since sequence length serves as a practical proxy for information quantity, we formulate a normalization loss based on the prefix length ($l_p$) and full sequence length ($l_g$):
\vspace*{-0.5\baselineskip} 
\begin{equation}
L_\mathrm{norm} = \mathrm{SmoothL1Loss}\left(\overline{p},~ \frac{\min(l_p, l_b) \times 0.5 + (l_g - l_p)}{l_g}\right)
\end{equation}
where $\overline{p}$ represents the sequence's mean gate output, and $l_b$ is a buffer hyperparameter (set to 1.5 seconds in our experiments) that prevents abrupt score changes near prefix boundaries. The 0.5 weighting factor accounts for the partial information available at prefix edges.

\paragraph{Pre-Sigmoid Gaussian Noise} Following prior work \cite{raffel2017Online}, we incorporate Gaussian noise before Sigmoid activation to promote discretization of gate output. While this discretization does not directly improve model performance, it significantly enhances robustness by reducing sensitivity to threshold selection across diverse tasks. In our implementation, we apply zero-mean unit-variance Gaussian noise ($\sigma_R=1$) to the pre-Sigmoid logits.

Combined with our normalization technique, this approach enables precise control over both the mean and variance of gating scores. The complete training objective is:
\begin{equation}
\mathcal{L}^\mathrm{total} = \mathcal{L}^\mathrm{simul} + w_\mathrm{n} \cdot \mathcal{L}^\mathrm{norm}
\end{equation}
where we set a small normalization weight $w_\mathrm{n}=0.01$.

\subsection{Streaming TTS}\label{sec:tts}
A common practice in speech-to-speech translation (S2ST) systems involves integrating a speech-to-text translation (S2TT) module with a text-to-speech (TTS) component. For simultaneous translation scenarios, however, a streaming TTS system is essential to preserve the low-latency requirement of the overall pipeline. While specialized streaming TTS systems exist, we demonstrate that SimulMEGA offers a universal and straightforward method to convert a standard transformer-based autoregressive TTS system into a streaming-capable variant.

For our experiments, we adopt CosyVoice 2 as the base model, which consists of a decoder-only language model and a flow-based acoustic model. During adaptation, we retain the flow model and fine-tune the language model using the techniques and loss functions outlined in Sections 3.2 and 3.3. Randomly initialized routing gate and the MoE refiner are incorporated during training. Since the backbone lacks an encoder, we derive the prefix and offline hidden states ($h^\mathrm{prefix}$ and $h^\mathrm{offline}$) from the final hidden representations of the text tokens. A detailed architectural diagram is provided in the appendix. We compare our approach against CosyVoice2’s default dual-stream mode.

Through this work, we not only develop a more comprehensive S2ST system capable of real-time speech-to-speech translation but also validate the compatibility of SimulMEGA with diverse tasks and model architectures. We leave an end-to-end S2ST system with SimulMEGA for future work.

%% file: appendix.tex
\newpage

\appendix

\section{SimulMEGA-TTS Implementation Details}

\begin{figure}[htp]
    \centering
    \includegraphics[width=1.0\linewidth]{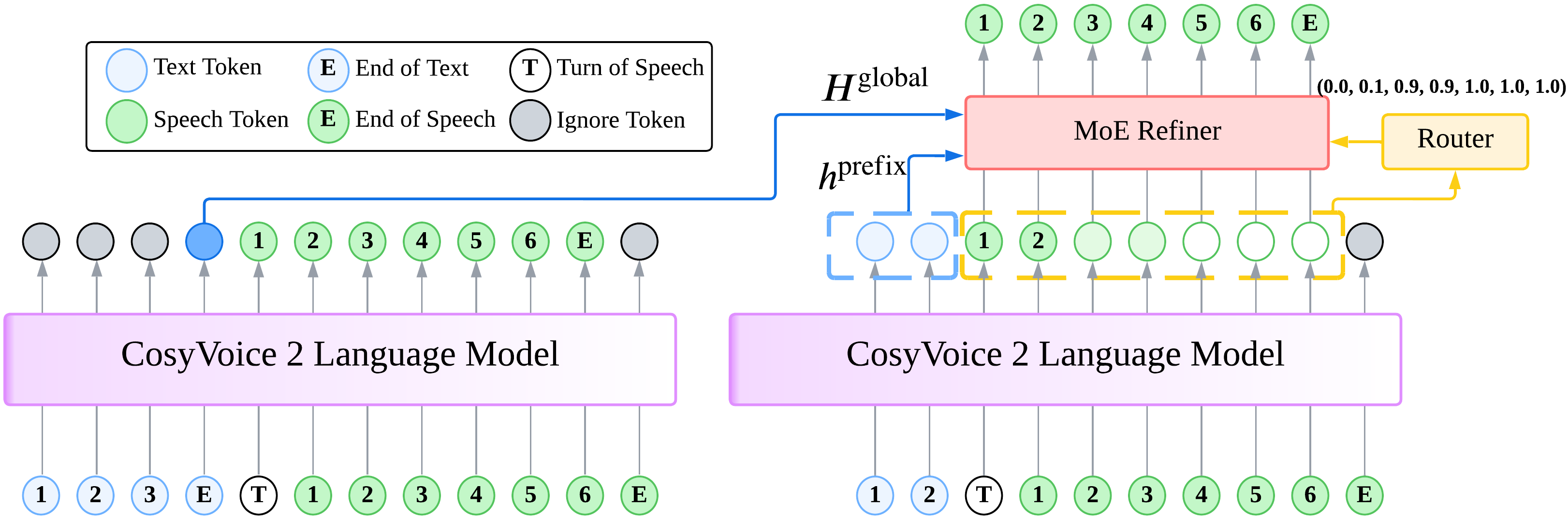}
    \caption{Illustration of SimulMEGA-TTS. The left uses offline text input while the right uses prefix text input for policy learning. }
    \label{fig:tts}
\end{figure}

Figure~\ref{fig:tts} shows how to implement SimulMEGA on top of a uni-directional decoder-only backbone. We take CosyVoice 2 as the base model and fine-tune it into SimulMEGA-TTS. We take the last layer hidden states output of text tokens as the $h^\mathrm{prefix}$ and the hidden state of the text EOS token as $H^\mathrm{global}$. 

In S2TT, normally, when the input reaches the end of the stream, we no longer turn to the read/write strategy and instead continue the generation until the EOS token appears. However, in the TTS task, the end of speech is relatively ambiguous. Therefore, if we employ the same strategy as S2TT, the hallucination problem may occur at the end of the generation. Therefore, we employ a mixed strategy to tell the end of the generation:
\begin{equation}
\text{End of Generation} = \begin{cases}
\text{True} & \text{if } E_\mathrm{t} \in T_\mathrm{text} \text{ and } (E_s \in T_\mathrm{speech} \text{ or } p_{\cdot,i}>\lambda_\mathrm{end}) \\
\text{False} & \text{otherwise}
\end{cases}
\end{equation}
Where $E_\mathrm{t/s}$ denotes text/speech EOS token. $T_\mathrm{text/speech}$ denotes text or speech tokens in the current sequence. $p_{\cdot, i}$ is the router output at the current position after all input streams are read. $\lambda_\mathrm{end}=0.9$.

\section{Experiment Details}
\subsection{S2TT Baselines}
\begin{enumerate}[leftmargin=0.5cm]
    \item \textbf{Meta Seamless series:} Meta Seamless series contains a set of models including SeamlessM4T-large-v2 for offline translation, SeamlessStreaming for simultaneous translation and SeamlessExpressive for voice cloning. They also combines the capability of SeamlessStreaming with SeamlessExpressive into a single model named Seamless. In our experiment we use SeamlessM4T-large-v2 for offline evaluation, SeamlessStreaming for simultaneous S2TT evaluation and Seamless for simultaneous S2ST evaluation.
    \item \textbf{Wait-K:} For each input test sample, We use a VAD to determine the starting time. Then we wait for two chunks (1.28s) and start generation at a fixed ratio, ranging from 0.5 to 2.5 tokens per chunk.
    \item \textbf{DiG-SST:} Following the original paper, we freeze the base model and add three learnable policy layers on top of the text decoder to predict the divergence between prefix and global input. It was trained for 100k steps on the same training set of SimulMEGA. 
    \item \textbf{EDATT:} We follow the paper and official code\footnote{\url{https://github.com/hlt-mt/FBK-fairseq/blob/master/examples/speech_to_text/simultaneous_translation/agents/v1_1/simul_offline_edatt.py}} for implementation. We re-tuned the thresholds and hyperparameter and set attention layer to be the 8th layer and set attention frame to be 4.
    \item \textbf{AlignATT:} We follow the updated official code\footnote{\url{https://github.com/hlt-mt/FBK-fairseq/blob/master/examples/speech_to_text/simultaneous_translation/agents/v1_1/simul_alignatt_seamlessm4t.py}}. A frame-wise normalization is applied to the attention score to mitigate maximum attention fixation. 
\end{enumerate}

In our experiment we observed that both DiG-SST and AlignATT behave poorly on all X2ZH pairs, which hinders the overall performance. We hypothesis that this might be caused by the UTF-8 byte-level BPE tokenizer. It encodes each CJK character into multiple tokens, potentially leading to unexpected output distribution and attention pattern.

\subsection{Quality-latency variable}
Table \ref{tab:var} shows the choice of the variable that controls the quality-latency tradeoff in the main experiment. From left to right the quality and latency gradually increase. 
\begin{table}[!htb]
    \centering
    \caption{Quality-latency variables in the main experiment.}
    \label{tab:var}
    \begin{tabular}{cccc}
    \toprule
        Method & variable & S2TT & S2ST \\ \midrule
        SimulMEGA & $\lambda$ & [0.9, 0.7, 0.5, 0.3, 0.1] & [0.8, 0.7,... , 0.2] \\ 
        Seamless & $t_{EMMA}$ & [0.3, 0.5, 0.7, 0.9, 1] & [0.3, 0.5, 0.7, 0.9] \\ 
        DiG-SST & $\lambda$ & [0.1, 0.07, 0.05, 0.03, 0.01] & - \\ 
        EDATT & $\alpha$ & [0.05, 0.03, 0.02, 0.015, 0.01] & - \\ 
        AlignATT & $f$ & [32, 48, 64, 80, 96] & - \\ 
        Wait-K & ratio & [2.5, 2.0, 1.5, 1.0, 0.5] & - \\ 
        \bottomrule
    \end{tabular}
\end{table}

\section{S2TT computation overhead.}
\begin{figure}[htp]
    \centering
    \includegraphics[width=0.8\linewidth]{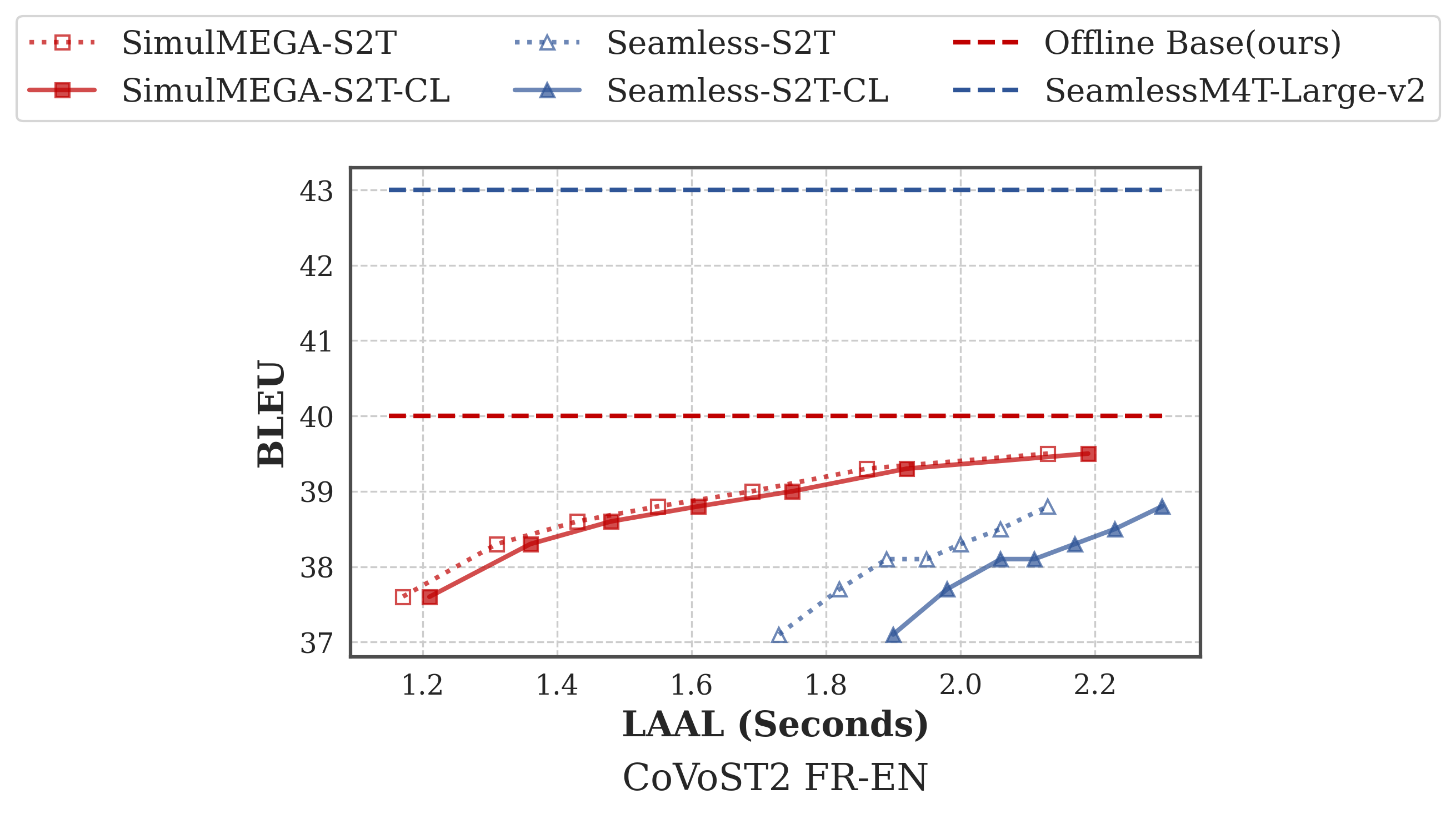}
    \caption{Computation aware quality-latency curve on CoVoST2 FR-EN. CL denotes Computation-Aware.}
    \label{fig:cl}
\end{figure}

For evaluating the computation overhead, we plot the computation-aware(CL) quality-latency trade-off curve in the CoVoST2 FR-EN testset. Computation-aware latency considers the actual inference time of the model. The evaluation is performed on a fully idle machine with a single Nvidia-H100 GPU. As shown in Figure~\ref{fig:cl}, for SimulMEGA, the extra AL due to computation is around 50 ms, whereas the number for Seamless is around 200 ms. 

\section{S2ST System Deployment \& Latency.}
We deploy our S2ST system on an Ubuntu server equipped with NVIDIA H100 GPUs. For the frontend and backend, we utilize FastRTC\footnote{\url{https://github.com/gradio-app/fastrtc}} and FastAPI\footnote{\url{https://github.com/fastapi/fastapi}}, respectively. To enhance inference efficiency, vLLM\footnote{\url{https://github.com/vllm-project/vllm}} is used to accelerate the language model in TTS, and TensorRT\footnote{\url{https://github.com/NVIDIA/TensorRT}} is used to accelerate the flow model.

We evaluate the system latency with streaming inputs, using four language pairs (ZH-EN, EN-ZH, FR-EN, FR-ZH). For each direction, we test two samples, each 20–30 seconds in length, and record event timestamps for latency calculation. Virtual Audio Cable\footnote{\url{https://vac.muzychenko.net/en/}} software is employed to eliminate echo interference and ensure consistent audio input across tests. The average latency results are presented in Table~\ref{tab:latency}. Here, the S2T/S2S Start Offset indicates the time required to generate the first text or speech chunk, while the S2T/S2S End Offset measures the delay of the final text chunk or speech frame relative to the end of the source speech.

It should be noted that these results provide only a rough estimation of system latency, as actual values may vary due to server or network conditions, output content length, speech rate, or other factors.

\begin{table}[htp]
    \centering
    \caption{Simultaneous S2ST system latency statistics (in seconds).  N denotes using non-stream TTS and S denotes stream TTS}
    \label{tab:latency}
    \begin{tabular}{llcccc}
    \toprule
         && Offline&\multicolumn{2}{c}{SimulMEGA}&Seamless\\
         &&  N&  N& S&-\\
    \midrule
          S2TT&Start Offset&  13.56&  3.12& 2.95&3.55\\
          &End Offset&  0.78&  0.44& 0.58&0.61\\
          \midrule
  S2ST&Start Offset& 15.98& 7.43& 3.87&4.33\\
          &End Offset&  14.16&  7.65&  7.54&6.42\\
    \bottomrule
    \end{tabular}
\end{table}

\section{Numerical results}
Numerical results of simulMEGA S2TT on CoVoST 2 testset is shown in Table \ref{tab:num_s2t}.
\begin{table}[!h]
    \centering
    \caption{Numerical results of simulMEGA S2TT on CoVoST 2 testset.}
    \label{tab:num_s2t}
    \begin{tabular}{cccccccc}
    \toprule
        \multicolumn{2}{c}{\textbf{Threshold}} & \textbf{0.9} & \textbf{0.7} & \textbf{0.5} & \textbf{0.3} & \textbf{0.1} & \textbf{0} \\ 
        \midrule
         \multirow{2}{*}{zh-en} & BLEU & 21.06 & 25.37 & 26.01 & 26.40 & 26.63  & 26.89  \\ 
         & LAAL & 1.395 & 2.249 & 2.539 & 2.807 & 3.180  & - \\ 
         \midrule
        \multirow{2}{*}{de-en} & BLEU & 32.90 & 36.08 & 36.94 & 37.15 & 37.41  & 37.44  \\ 
         & LAAL & 1.217 & 1.862 & 2.149 & 2.420 & 2.856  & - \\ 
         \midrule
        \multirow{2}{*}{es-en} & BLEU & 37.21 & 40.27 & 40.94 & 41.28 & 41.46  & 41.51  \\ 
         & LAAL & 1.047 & 1.521 & 1.760 & 2.001 & 2.405  & - \\ 
         \midrule
        \multirow{2}{*}{fr-en} & BLEU & 36.50 & 38.76 & 39.33 & 39.60 & 39.76  & 40.02  \\ 
         & LAAL & 0.952 & 1.391 & 1.590 & 1.812 & 2.195  & - \\ 
         \midrule
        \multirow{2}{*}{it-en} & BLEU & 34.64 & 37.42 & 38.12 & 38.57 & 38.89  & 38.90  \\ 
         & LAAL & 1.101 & 1.621 & 1.873 & 2.125 & 2.512  & - \\ 
         \midrule
        \multirow{2}{*}{en-zh} & BLEU & 39.93 & 42.77 & 43.11 & 43.17 & 43.42  & 43.57  \\ 
         & LAAL & 1.539 & 2.305 & 2.680 & 2.976 & 3.381  & - \\ 
         \midrule
        \multirow{2}{*}{en-de} & BLEU & 29.65 & 32.89 & 33.16 & 33.61 & 33.66  & 33.75  \\ 
         & LAAL & 1.266 & 1.813 & 2.121 & 2.404 & 2.808  & - \\ 
    \bottomrule
    \end{tabular}
\end{table}

\section{S2ST Case Study}
We visualize two S2ST example of SimulMEGA in FR-EN and ZH-EN, respectively. The second row is the isochronic text label of source speech and the forth row indicates the timestamps when each target text chunk are generated.

\begin{figure}[htp]
    \centering
    \includegraphics[width=1.0\linewidth]{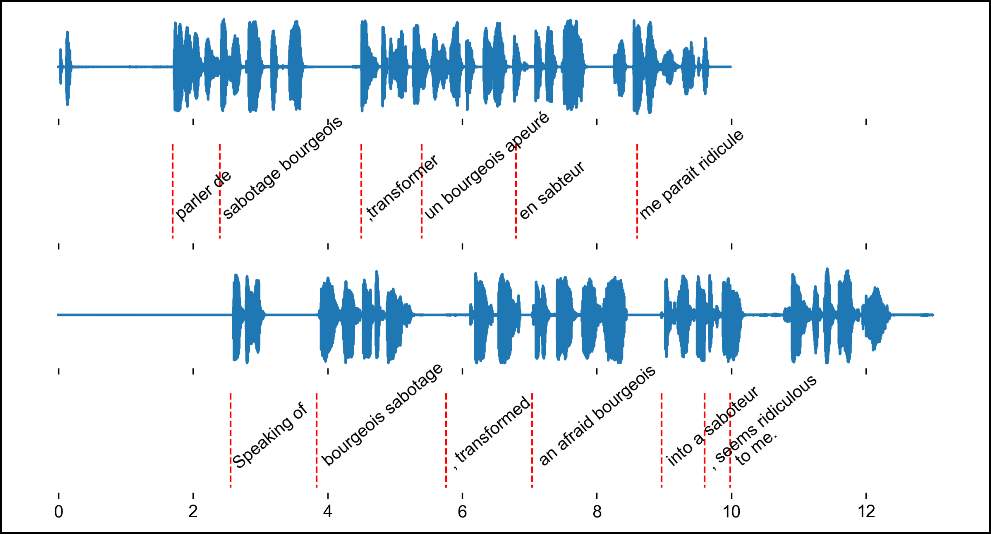}
    \caption{(a) FR-EN}
\end{figure}

\begin{figure}[htp]\ContinuedFloat
    \centering
    \includegraphics[width=1.0\linewidth]{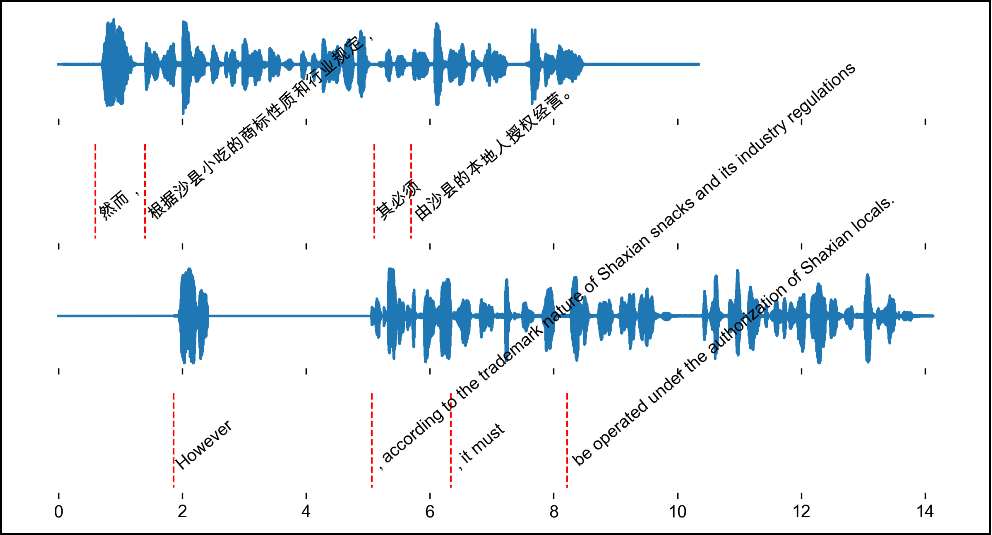}
    \caption{(b) ZH-EN}
\end{figure}